\documentclass[letterpaper, 10 pt, conference]{ieeeconf}  

\IEEEoverridecommandlockouts                              

\usepackage[caption=false]{subfig}
\usepackage{url}
\usepackage[acronym]{glossaries}
\usepackage{multirow}
\usepackage{makecell}
\usepackage{placeins}
\usepackage{xcolor}
\usepackage{supertabular,booktabs}
\usepackage{tabularx}
\usepackage{balance}
\usepackage[colorinlistoftodos,prependcaption,textsize=tiny]{todonotes}
\newcolumntype{Y}{>{\centering\arraybackslash}X}

\usepackage{tikz}
\usepackage{textcomp}
\usepackage{hyperref}
\usepackage{lipsum}

\newcommand\copyrighttext{%
  \footnotesize \textcopyright 2019 IEEE. Personal use of this material is permitted.
  Permission from IEEE must be obtained for all other uses, in any current or future
  media, including reprinting/republishing this material for advertising or promotional
  purposes, creating new collective works, for resale or redistribution to servers or
  lists, or reuse of any copyrighted component of this work in other works.
  }
\newcommand\copyrightnotice{%
\begin{tikzpicture}[remember picture,overlay]
\node[anchor=south,yshift=10pt] at (current page.south) {\fbox{\parbox{\dimexpr\textwidth-\fboxsep-\fboxrule\relax}{\copyrighttext}}};
\end{tikzpicture}%
}

\title{\LARGE \bf
Masking Salient Object Detection, a Mask Region-based Convolutional Neural Network Analysis for Segmentation of Salient Objects}

\author{
        Bruno A. Krinski, Daniel V. Ruiz, Guilherme Z. Machado and Eduardo Todt \\
        Department of Informatics, Federal University of Paran\'a (UFPR), Curitiba, PR, Brazil \\ 
         \textit{\{bakrinski, dvruiz, gzmachado, todt\}@inf.ufpr.br}
        } 

\begin{document}

\maketitle
\copyrightnotice
\thispagestyle{empty}
\pagestyle{empty}

\newacronym{mae}{MAE}{Mean Absolute Error}
\newacronym{resnet}{ResNet}{Residual Network}
\newacronym{svm}{SVM}{Support Vector Machine}
\newacronym{fpn}{FPN}{Feature Pyramid Network}
\newacronym{rpn}{RPN}{Region Proposal Network}
\newacronym{sod}{SOD}{Salient Object Detection}
\newacronym{fcn}{FCN}{Fully Convolutional Network}
\newacronym{cnn}{CNN}{Convolutional Neural Network}
\newacronym{hed}{HED}{Holistically-Nested Edge Detector}
\newacronym{slic}{SLIC}{Simple Linear Iterative Clustering}
\newacronym{roialign}{RoIAlign}{Region of Interest Alignment}
\newacronym{maskrcnn}{Mask-RCNN}{Mask Region-based Convolutional Neural Network}
\newacronym{fasterrcnn}{Faster-RCNN}{Faster Region-based Convolutional Neural Network}

\begin{abstract}

In this paper, we propose a broad comparison between \glspl*{fcn} and \glspl*{maskrcnn} applied in the \gls*{sod}
context. Studies in the \gls*{sod} literature usually explore architectures based in \glspl*{fcn} to detect salient
regions and objects in visual scenes. However, besides the promising results achieved, \glspl*{fcn} showed issues in
some challenging scenarios. Fairly recently studies in the \gls*{sod} literature proposed the use of a \gls*{maskrcnn} approach to overcome
such issues. However, there is no extensive comparison between the two networks in the \gls*{sod} literature
endorsing the effectiveness of \glspl*{maskrcnn} over \gls*{fcn} when segmenting salient objects. Aiming to
effectively show the superiority of \glspl*{maskrcnn} over \glspl*{fcn} in the \gls*{sod} context, we compare
two variations of \glspl*{maskrcnn} with two variations of \glspl*{fcn} in eight datasets widely used in the literature
and in four metrics. Our findings show that in this context \glspl*{maskrcnn} achieved an improvement on the F-measure
up to 47\% over \glspl*{fcn}.

\end{abstract}


\section{Introduction}
\label{sec:introduction}
Visual Salience (or Visual Saliency) is the characteristic of some objects or regions in images to stand out from their surrounding
elements, attracting the attention of the human visual system~\cite{itti:visual_salience}. Finding Visual Saliences has a wide range
of applications in Computer Vision and Image Processing tasks like detecting, recognizing and tracking objects, image cropping,
image resizing, and video compression or summarization~\cite{lee:deep_saliency,xi:a_fast_and_compact}. In the robotics field, finding
salience regions in images can be used to improve algorithms of location and mapping of environments~\cite{todt:outdoor_landmak_view} and
object localization~\cite{robs1,robs2}. The \glsreset{sod}\gls*{sod} research field aims to detect and segment the regions in images with the
high probability of being salient. 

\begin{figure}[!htb]
	\centering
	\captionsetup[subfigure]{width=0.45\linewidth}
	\subfloat[][Example of image containing one salient object.]{
	    \includegraphics[width=0.45\linewidth]{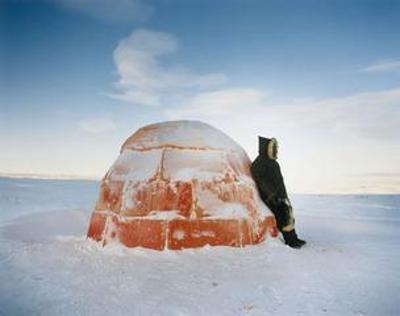}
	    \label{fig:image_example}
	}
	\subfloat[][The respective Saliency Map (the ground truth).]{
	    \includegraphics[width=0.45\linewidth]{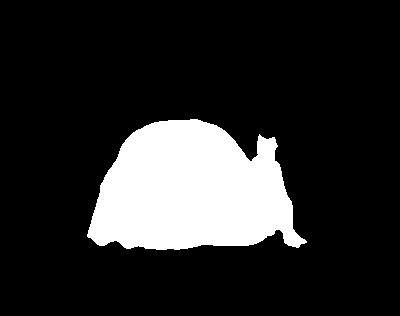}
	    \label{fig:mask_example}
	}
	
	\subfloat[][The Saliency Map generated by the \gls*{fcn} (in green).]{
	    \includegraphics[width=0.45\linewidth]{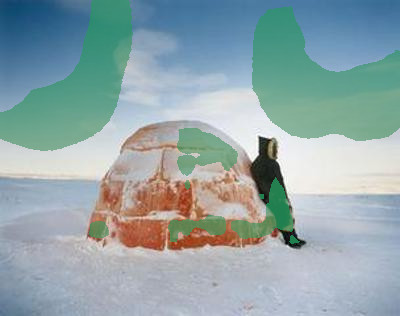}
	    \label{fig:fcn_example}
	}
	\subfloat[][The Saliency Map generated by the \gls*{maskrcnn} (in green).]{
	    \includegraphics[width=0.45\linewidth]{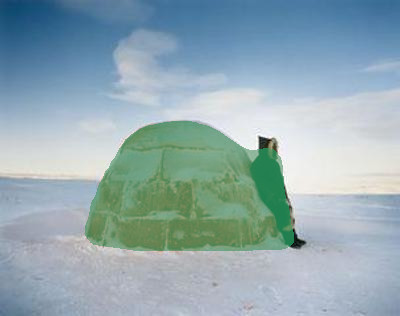}
	    \label{fig:mask_example2}
	}
	\caption[]{Comparison between \gls*{fcn} and \gls*{maskrcnn} segmentation. The \gls*{fcn} found the salient object
	(\ref{fig:image_example}). However, the \gls*{fcn} generated a very poor Saliency Map (\ref{fig:fcn_example}). In~(\ref{fig:mask_example2}), the
	\gls*{maskrcnn} generated a Saliency Map much closer to the expected ground-truth (\ref{fig:mask_example}) and drastically reduced the false negative
	error. The illustrative sub-figures belong to the DUT-OMRON dataset.}
	\label{fig:segmentation_example}    
\end{figure}

Recent works in \gls*{sod} literature~\cite{xi:deep_saliency,tang:deeply_supervised,zhang:deep_salient,tang:saliency_detection,xi:a_fast_and_compact,hou:deeply_spervised,xie:holistically,zhang:reflection,zhang:reflection2,li:weakly} proposes the use of 
architectures based on the \glsreset{fcn}\gls*{fcn} proposed by Long \textit{et al.}~\cite{long:fcn}. However, despite the promising
results achieved by \glspl*{fcn}, there are relevant inaccuracies in the segmentation result, as presented in 
Fig.~\ref{fig:segmentation_example}. Trying to solve similar problems, Nguyen~\textit{et al.}~\cite{spasod} fairly recently proposed
the use of a \glsreset{maskrcnn}\gls*{maskrcnn}~\cite{girshick:maskrcnn} approach to \gls*{sod}. However an extensive comparison
between the \gls*{fcn} and the \gls*{maskrcnn} was not performed.

Taking in mind the shallow evaluation between the \gls*{fcn} and the \gls*{maskrcnn} available in the literature, in this work we
propose a broad evaluation between the two early mentioned segmentation approaches. To perform the evaluation we propose four
\gls*{sod} approaches, two based on \gls*{maskrcnn}, and another two based on \gls*{fcn}. The comparison was made in eight well-known
\gls*{sod} datasets, by using the F-measure, Precision, Recall and \gls{mae} metrics. The obtained results show the superiority of
the \gls*{maskrcnn} over the \gls*{fcn} approaches from up to 47\% of F-measure.
\section{Related Works}
The approaches proposed in the \gls*{sod} literature can be divided into three waves~\cite{borji:salienci_survay}:
handcrafted-based, \gls*{cnn}-based~\cite{lecun:cnn}, and \gls*{fcn}-based methods. The first attempts to find saliency regions in
images were based on handcrafted features and addressed the problem through heuristics to find the saliency regions and traditional
Computer Vision techniques to perform the segmentation of the saliency regions. 

One of the earliest studies presented in the \gls*{sod} literature was proposed by Itti \emph{et 
al.}~\cite{itti:a_model_of_saliency} in 1998 and started a wave of works on the subject. In the proposed method, a 
center-surrounding technique is utilized to extract information of color, intensity, and orientation from images. In the last step,
they generate a set of feature maps for each extracted feature, which are combined to generate the Saliency Map.

Achanta \emph{et al.}~\cite{achanta:salient_reagion_detection} extracted patches from the images to generate a feature vector of
color and luminance. Then, a window surrounding approach is applied to calculate the difference of contrast between each pixel and
its neighbors. Jiang \emph{et al.}~\cite{jiang:automatic_salient} combined context and shape information from the image to produce
the Saliency Map. In the context extraction step, the image is segmented in super-pixels, and the saliency of a determined 
super-pixel is calculated as a function of its spatial neighbors. In the shape extraction step, they use an edge detector algorithm
to find the contours of the objects.

Despite the promising results achieved by the handcrafted based methods, they lacked generalization capacity, causing decay in the
algorithm performance in challenging scenarios. In order to overcome such limitations, \gls*{cnn}, an approach
already applied in other Computer Vision problems like object detection and image classification started to be widespread in the
\gls*{sod} literature. When the \glspl*{cnn} started to be applied in the \gls*{sod}, the common strategy presented in the 
literature (decompose an input image in patches or super-pixels) continued to be used similarly to handcrafted
approaches~\cite{borji:salienci_survay}.

One of the earliest models based on \glspl*{cnn} was presented by He \emph{et al.}~\cite{He:supercnn} (SuperCNN) in 2015. The
proposed method decomposed each super-pixel in two contrast sequences of color uniqueness and color distribution. Each sequence
feeds a different \gls*{cnn} with 1D convolutions in parallel. Wang \emph{et al.} applied the \gls*{fasterrcnn}~\cite{girshick:fasterrcnn} architecture to classify the
super-pixels as being part of the saliency or background regions. Low-level features (contrast and backgroundness) and an edge-based
propagation method are applied to refine the Saliency Map.

Fusing local-context information to detect high-frequency content and global-context information to suppress homogeneous regions
through holistic contrast and color statistics from the entire image, is another approach explored by \gls*{cnn}-based
methods~\cite{wang:deep_networks}. An example is the work of Zhao \emph{et al.}~\cite{zhao:saliency_detection}, that proposed a
method with a Neural Network to find the global-context saliences (upper-branch) and another Neural Network to extract local-context
saliences (lower-branch). 

Recent studies in the \gls*{sod} literature started to explore \gls*{fcn}~\cite{long:fcn}, an improvement of the standard 
\glspl*{cnn} developed for Semantic Segmentation problems. The \gls*{fcn} performs a more accurate segmentation in a per-pixel level
and overcomes problems caused by fully connected layers such as blurriness and incorrect predictions near the boundaries of salient
objects in super-pixel based methods~\cite{borji:salienci_survay}. 

Zhang \emph{et al.}~\cite{zhang:deep_salient} adapted the \gls*{resnet}-101 network to perform segmentation. In a second step, the
\gls*{slic} algorithm is applied in the input image to generate super-pixels of three different scales. The Saliency Map produced by
the \gls*{fcn} is utilized to assign a saliency value to each super-pixel. An energy minimization function is applied to refine
the saliency value of each super-pixel. In the last step, the Saliency Maps at each scale are utilized, generating the final
Saliency Map.

Tang \emph{et al.}~\cite{tang:saliency_detection} constructed the Saliency Map by fusing the Saliency Map of a \gls*{fcn} with the
Saliency Map of a \gls*{cnn}. The \gls*{fcn} performs a pixel-level segmentation while the \gls*{cnn} performs super-pixel level
segmentation, generating two Saliency Maps, which are concatenated to generate the Saliency Map. Xi \emph{et 
al.}~\cite{xi:a_fast_and_compact} proposed an end-to-end network with three sequential steps. A VGG-16 network is utilized in the
first step to extract feature maps. Three fully convolutional layers are added at the end of the VGG-16 in the second step to make a
non-linear regression. In the last step, bicubic interpolation is applied to restore the size of the Saliency Map.

Hou \emph{et al.}~\cite{hou:deeply_spervised} proposed a top-down approach to combine low-level with high-level features. A
modification of the \gls*{hed}~\cite{xie:holistically} is utilized to adapt the VGG-16 network for edge and boundary detection of
salient objects. The output of each block in the network is concatenated with the output of all previews blocks through the short
connections, generating a segmentation mask for  each block of the network. In the last step, they concatenated all segmentation
masks with a weighted fusion layer.

\begin{figure*}[!htb]
    \centering
    \includegraphics[width=0.90\linewidth]{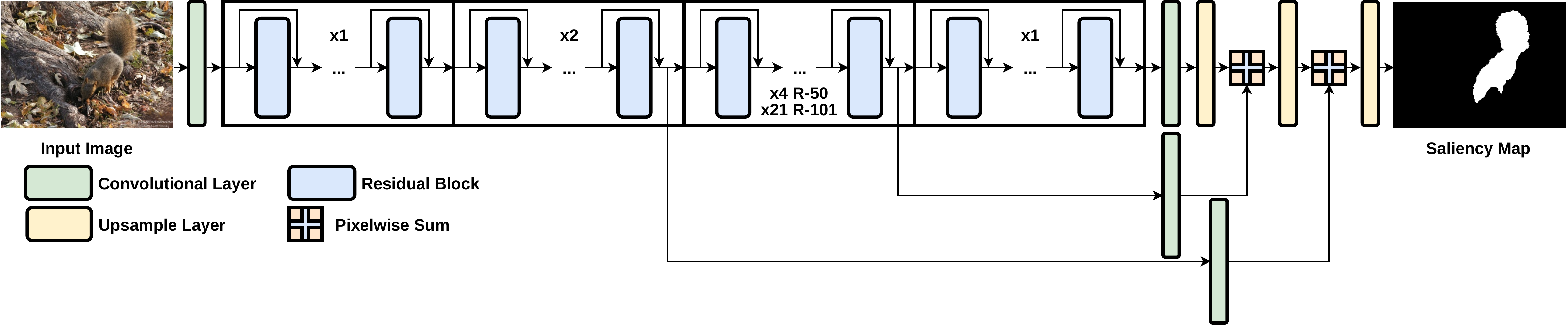}
    \caption{The \gls*{resnet} architectures converted to \gls*{fcn}-8 in the KittiSeg framework. Each residual block (represented in
    blue) is composed of three convolutional layers in sequence with filters of size $3\times3$, $1\times1$, and $3\times3$, 
    respectively. The input of each residual block is added to its output through the skip connection. R-50 represent 
    \gls*{resnet}-50 and R-101 represent \gls*{resnet}-101. The output of the network is the Saliency Map.}
    \label{fig:resnetfcn}
\end{figure*}

Nguyen \emph{et al.}~\cite{spasod} proposed a framework with two steps to segment salient objects. In the first step, they utilized
\glspl*{fcn} to generate a Saliency Map, called explicit Saliency Map, that targets the human common sense of salient objects and
focus in the learning of different objects previously defined as salient. In the second step, they utilized an approach based in
super-pixels to generate a second Saliency Map, called implicit Saliency Map, that focus in the learning of salient objects not
belonging to the list of salient objects defined in the first step. Then, they defined a function that fuses both Saliency Maps.
In the first step, they utilized four variations of \gls*{fcn} and also applied a \gls*{maskrcnn} to generate the explicit Saliency 
Map. However, the authors did not performed an explicit comparison between the \gls*{fcn} and \gls*{maskrcnn}. They only compared
the entire framework proposed, changing the network utilized to generate the first saliency map. This brief comparison was performed
only in one dataset and the entire framework was evaluated only in three datasets.
 
\section{Evaluated Networks}
\subsection{Fully Convolutional Network (FCN)}
The \gls*{fcn} was designed by Long \emph{et al.}~\cite{long:fcn} in 2015 to perform Semantic Segmentation. In a \gls*{fcn}, the
fully connected layers of a \gls*{cnn} are converted in convolutional layers. Thus, the output of the network is a heatmap with dense
prediction values. In the end, bilinear interpolation is applied to convert the heatmap in a segmented image with the same size of
the input image.

Among the \glspl*{fcn} models proposed by Long \emph{et al.}~\cite{long:fcn}, the \gls*{fcn}-8 achieved better segmentation results
and was utilized in our work. In the convolutional step of \gls*{cnn}, there is an information loss as the image goes deeper into the
network and the \gls*{fcn}-8 attempt to recover information from the top layers of the network to help in the upsampling step. 

In order to evaluate the \gls*{fcn}, we use the KittiSeg~\cite{teichmann:kittiseg} open-source framework\footnote{https://github.com/MarvinTeichmann/KittiSeg}, following Bezerra \emph{et al.}~\cite{bezerra2018robust}, which converts two versions of \gls*{resnet}, with 50 and 101
layers, in \gls*{fcn}-8. Fig.~\ref{fig:resnetfcn} presents the \gls*{resnet}-50 and \gls*{resnet}-101 converted to \gls*{fcn}-8 in the
KittiSeg implementation. The \glspl*{resnet} are structure by a sequence of four building blocks, with each building block containing a sequence of residual blocks. The difference between each building block is the number of filters in the convolutional layers.

In the implementation utilized, the output of the last building block enters a convolutional and upsample layers, generating a 
feature map which is pixelwise added to the third building block. The resulting feature map suffers an upsample and is pixelwise 
added to the output of the second building block, generating a feature map which is upsampled to generate the final Saliency Map. The
output of the second and third building blocks also suffers a convolution operation before being added to the feature map.

\subsection{Mask Region-based Convolutional Network (Mask-RCNN)}
\label{sec:maskrcnn}
The \gls*{maskrcnn} was developed by Girshick \emph{et al.}~\cite{girshick:maskrcnn} for Instance Segmentation and is composed of two
sub-networks: the \gls*{fasterrcnn} to perform object detection and classification, and an \gls*{fcn} to perform object segmentation.
So, the \gls*{maskrcnn} has three outputs: the classes of the objects in the image, the coordinates of the bounding boxes that
surround these objects and the segmentation mask.

\begin{figure*}[!htb]
    \centering
    \includegraphics[width=0.90\linewidth]{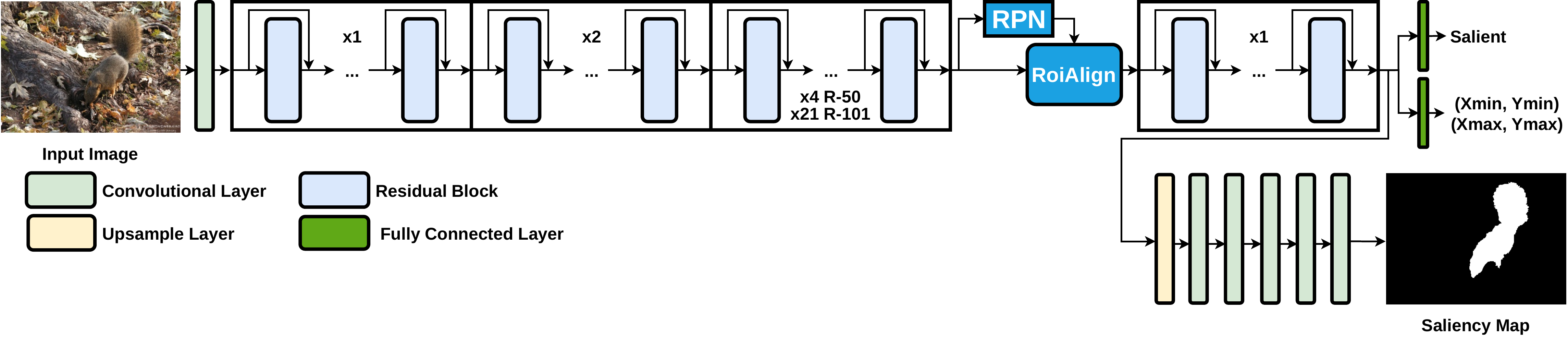}
    \caption{The \gls*{resnet} architectures converted to \gls*{maskrcnn} in the Object Detection API. Each residual block
    (represented in blue) is composed of three convolutional layers in sequence with filters of size $3\times3$, $1\times1$, and
    $3\times3$, respectively. The input of each residual block is added to its output through the skip connection. R-50 represent
    \gls*{resnet}-50 and R-101 represent \gls*{resnet}-101. Between the third and fourth blocks, the \gls*{maskrcnn} has the
    \gls*{rpn} to generate a set of region proposals and the \gls*{roialign} to perform a max-pooling operation to convert all
    proposals to a same size feature map. The output of the network is the object classification (Salient), the bounding boxes
    coordinates, and the Saliency Map.}
    \label{fig:resnetmaskrcnn}
\end{figure*}

In order to evaluate the \gls*{maskrcnn}, we utilize the Object Detection API~\cite{huand:seed_accuracy}, an open-source 
framework \footnote{https://github.com/tensorflow/models/tree/master/research/object\_detection} which converts two versions of
\gls*{resnet}, with 50 and 101 layers in \glspl*{maskrcnn}. The implementation proposed in the Object Detection API converts
\glspl*{fasterrcnn} by adding a segmentation module attached in parallel with the classification and detection outputs.
Fig.~\ref{fig:resnetmaskrcnn} presents the \gls*{resnet}-50 and \gls*{resnet}-101 converted to \gls*{maskrcnn} in the Object
Detection API implementation. 

In the implementation utilized, the \gls*{maskrcnn} is structured as follows: three building blocks of the \gls*{resnet} are utilized
to generate a feature map. Then, a module called \gls*{rpn} is linked in the network to generate a set of region proposals in the
image with a high probability of being objects. These regions enter the \gls*{roialign} layer to generate feature maps with a fixed
size.

The region proposals enter the last building block of the \gls*{resnet}, generating a feature map which enters in two fully
connected layers attached in parallel to perform the classification and to adjust the bounding boxes. Also in parallel, the feature
map enter in a upsample layer followed by five convolutional layers to generate the Saliency Map.

\section{Experiments}
\subsection{Datasets}
In order to evaluate the network models, we trained the models on the MSRA10K~\cite{cheng:global_contrast} dataset and tested on
eight public datasets of salient objects widely used in the \gls*{sod} literature: DUT-OMRON~\cite{ruan:dutomron}, 
ECSSD~\cite{yan:ecssd}, HKU-IS~\cite{li:visual_saliency}, ICOSEG~\cite{icoseg}, PASCAL-S~\cite{li:pascals}, 
SED1~\cite{borji:benchmark}, SED2~\cite{borji:benchmark}, and THUR~\cite{ming:thur15k}. 

The datasets are composed by images containing salient objects with different biases (\emph{e.g.}, number of salient objects, image
clutter, center-bias) and the referent ground-truth mask with the expected segmentation of the salient objects. 

DUT-OMRON is a large dataset with 5,168 natural images with one or more salient objects and backgrounds relatively complex. ECSSD
contains 1,000 images with a great variety of scenarios and complex backgrounds. HKU-IS contains 4,447 challenging images with low
contrast or multi foreground objects. ICOSEG is an interactive co-segmentation dataset and contains 643 images with single or
multiple salient objects for each image. 

The PASCAL-S has 850 natural images with multiple complex objects and cluttered backgrounds and derives from the validation set of the PASCAL VOC 2012 segmentation challenge~\cite{pascal-voc-2012}. SED1 and SED2 are the smallest datasets, which contain only 100
images. The THUR dataset contains 6,232 images and is the unique dataset divided into classes, with a total of five classes
(butterfly, coffee mug, dog jump, giraffe, and plane).

Among all \gls*{sod} datasets utilized, the MSRA10K is the largest (10,000 images) and it is utilized to train the \gls*{fcn} and
\gls*{maskrcnn} models in our work, following similar work in the 
literature~\cite{nie:real_time_salient,tang:saliency_detection,zhang:deep_salient,xi:deep_saliency,xi:a_fast_and_compact,tang:deeply_supervised,babu:saliency_unified,zhang:reflection,zhang:reflection2}. In order to train the \gls*{maskrcnn}, it was necessary to
generate the bounding boxes of each salient object in the images of the MSRA10K dataset.

\subsection{Evaluation Metrics}
The architecture models trained in the \gls*{sod} problem are evaluated and compared through four metrics widely used in the
\gls*{sod} literature: Precision (Equation~\ref{eq:precision}), Recall (Equation~\ref{eq:recall}), F-measure 
(Equation~\ref{eq:fmeasure}), and \gls*{mae} (Equation~\ref{eq:mae}). 

\begin{figure*}[!htb]
	\centering
	\captionsetup[subfigure]{width=0.35\linewidth}
	\subfloat[][The learning curves with the F-measures of each fold generated by the \gls*{resnet}-50 converted to \gls*{fcn}.]{
	    \includegraphics[width=0.4\linewidth]{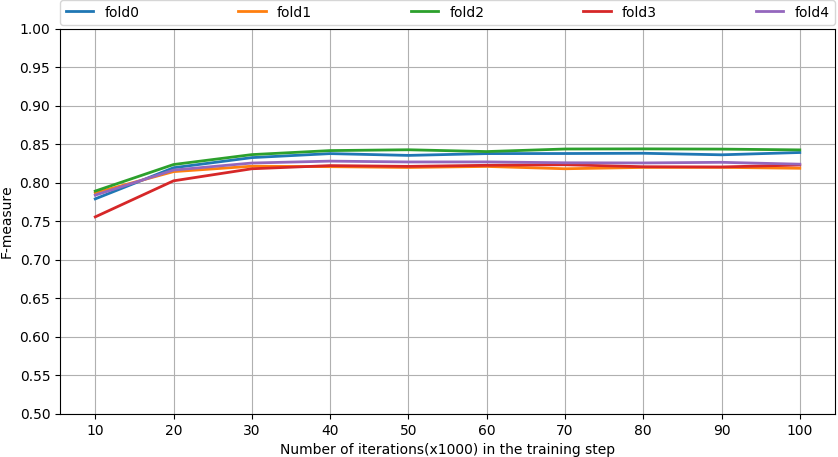}
	    \label{fig:resnet50_fcn_fmeasure}
	}
    \subfloat[][The learning curves with the F-measures of each fold generated by the \gls*{resnet}-101 converted to \gls*{fcn}.]{
	    \includegraphics[width=0.4\linewidth]{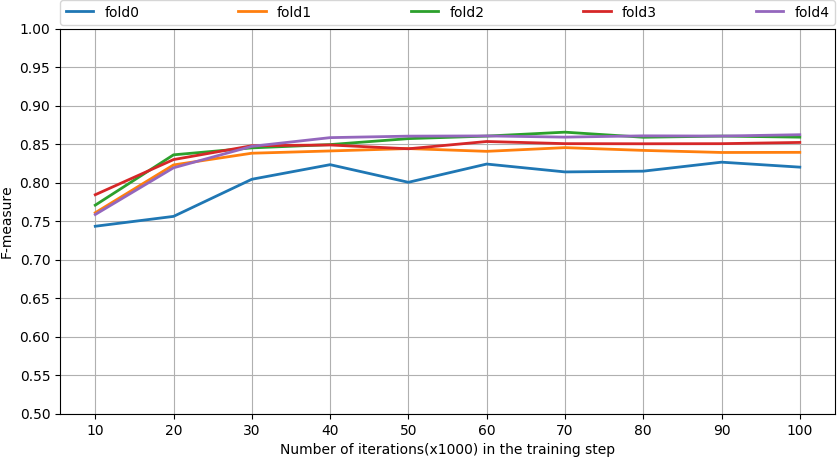}
	    \label{fig:resnet101_fcn_fmeasure}
	}
	
	\subfloat[][The learning curves with the F-measures of each fold generated by the \gls*{resnet}-50 converted to
	\gls*{maskrcnn}.]{
	    \includegraphics[width=0.4\linewidth]{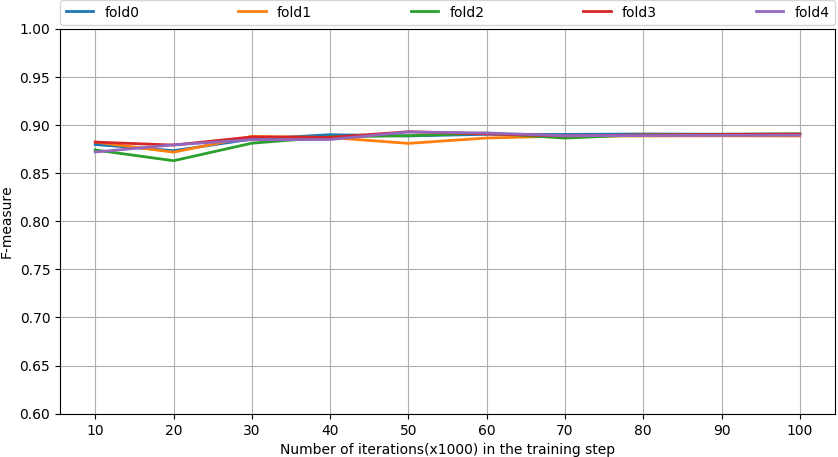}
	    \label{fig:resnet50_mask_fmeasure}
	}
    \subfloat[][The learning curves with the F-measures of each fold generated by the \gls*{resnet}-101 converted to 
    \gls*{maskrcnn}.]{
	    \includegraphics[width=0.4\linewidth]{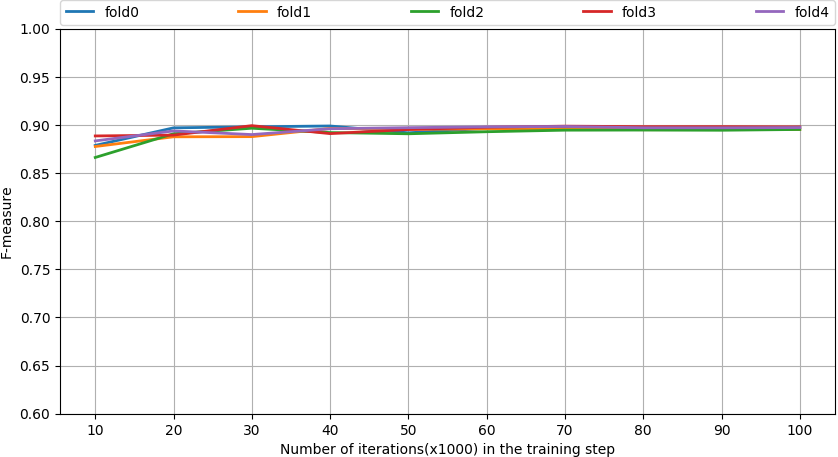}
	    \label{fig:resnet101_mask_fmeasure}
	}
    \caption[]{The learning curves with the F-measures of each fold in the validation step.}
    \label{validation}
\end{figure*}

\begin{equation}
    Precision = \frac{True Positive}{True Positive + False Positive}
\label{eq:precision}
\end{equation}
\begin{equation}
    Recall = \frac{True Positive}{True Positive + False Negative}
\label{eq:recall}
\end{equation}
\begin{equation}
    F-measure = \frac{(1+\beta^{2}) \times Precision \times Recall}{\beta^{2} \times Precision + Recall}
\label{eq:fmeasure}
\end{equation}
The $\beta$ in the F-measure formula changes the Precision and Recall importance. In the \gls*{sod} literature, 
$\beta^{2}$ receives the value $0.3$~\cite{borji:benchmark,achanta:frequency_tuned,cheng:global_contrast,perazzi:saliency_filters}
to increase the Precision importance.

The \gls*{mae} formula is presented in Equation~\ref{eq:mae}. The $W$ and $H$ are the image width and height, respectively. The $S$
is the predicted Saliency Map, and $G$ is the ground truth Saliency Map.

\begin{equation}
    MAE = \frac{1}{W \times H}\sum_{x=1}^{W}\sum_{y=1}^{H}|S(x,y)-G(x,y)|
\label{eq:mae}
\end{equation}
\subsection{Experiments}
The networks evaluated in our work were validated through a K-fold strategy with $k$ equal to five. For each fold, the networks were
trained with 100,000 iterations in the MSRA10K dataset. For each model evaluated, an initial learning rate was defined through 
preliminary experiments. 

The \glspl*{resnet} 50 and 101 converted to \gls*{fcn} received an initial learning rate of $10^{-6}$ and the \glspl*{resnet} 50 and
101 converted to \gls*{maskrcnn} received an initial learning rate of $10^{-4}$. An exponential learning rate decay, similar with
the proposed in the literature~\cite{learning_rate_decay}, was utilized and for all networks, the initial learning rate is divided
by 10 at the iteration 50,000 generating a learning rate that is divided by 10 again at the iteration 75,000.

Fig.~\ref{validation} presents the learning curves with the F-measures of each fold generated by the \gls*{resnet}-50 and 
\gls*{resnet}-101 converted to \gls*{fcn} and \gls*{maskrcnn}, respectively. After training the folds and choosing the best
iteration to train each model, we trained each network model five times using the entire MSRA10K dataset and tested in eight
public datasets widely utilized in the literature. Table~\ref{tab:tests} presents the average F-measure, Precision, Recall, and
\gls*{mae} achieved and the standard deviation of each model.

\begin{table*}
\centering
\caption{Comparison between \gls{fcn}-\gls{resnet}-50, \gls{fcn}-\gls{resnet}-101, \gls{maskrcnn}-\gls{resnet}-50, and 
\gls{maskrcnn}-\gls{resnet}-101. The best results with the \gls*{resnet}-50 backbone are highlighted in \textcolor{blue}{blue} and the best
results with the \gls*{resnet}-101 backbone are highlighted in \textcolor{red}{red}.} 
\resizebox{\textwidth}{!}{
\begin{tabular}{cccccccccccccccccc}

\toprule
\multicolumn{1}{c}{\textbf{Network}} &
\multicolumn{1}{c}{\textbf{Metric}} &
\multicolumn{2}{c}{\textbf{DUT-OMRON}} &
\multicolumn{2}{c}{\textbf{ECSSD}} &
\multicolumn{2}{c}{\textbf{HKU-IS}} &
\multicolumn{2}{c}{\textbf{ICOSEG}} &
\multicolumn{2}{c}{\textbf{PASCAL-S}} &
\multicolumn{2}{c}{\textbf{SED1}} &
\multicolumn{2}{c}{\textbf{SED2}} &
\multicolumn{2}{c}{\textbf{THUR}} 
\\ \midrule

\makecell{} &
\makecell{} &
\makecell{Avg} &
\makecell{Std} & 
\makecell{Avg} &
\makecell{Std} & 
\makecell{Avg} &
\makecell{Std} & 
\makecell{Avg} &
\makecell{Std} & 
\makecell{Avg} &
\makecell{Std} & 
\makecell{Avg} &
\makecell{Std} & 
\makecell{Avg} &
\makecell{Std} & 
\makecell{Avg} &
\makecell{Std} 
\\ \midrule

\makecell{\textbf{\gls*{fcn}} \\ \textbf{\gls*{resnet}-50}} & 
\makecell{\textbf{F-measure} \\ \textbf{Precision} \\ \textbf{Recall} \\ \textbf{\gls*{mae}}} &
\makecell{0.4999 \\ 0.6815 \\ 0.4411 \\ 0.1067} &
\makecell{0.0163 \\ 0.0172 \\ 0.0355 \\ 0.0015} &
\makecell{0.7902 \\ 0.8872 \\ 0.6975 \\ 0.0927} &
\makecell{0.0102 \\ 0.0169 \\ 0.0411 \\ 0.0054} &
\makecell{0.7618 \\ 0.8702 \\ 0.6659 \\ 0.0738} &
\makecell{0.0098 \\ 0.0183 \\ 0.0377 \\ 0.0031} &
\makecell{0.7488 \\ 0.8264 \\ 0.6926 \\ \textbf{\textcolor{blue}{0.0879}}} &
\makecell{0.0100 \\ 0.0170 \\ 0.0477 \\ 0.0072} &
\makecell{0.7141 \\ 0.8238 \\ 0.6466 \\ 0.1047} &
\makecell{0.0082 \\ 0.0159 \\ 0.0330 \\ 0.0034} &
\makecell{0.6410 \\ 0.8909 \\ 0.4971 \\ 0.1298} &
\makecell{0.0263 \\ 0.0132 \\ 0.0372 \\ 0.0090} &
\makecell{0.4083 \\ 0.8486 \\ 0.2646 \\ 0.1487} &
\makecell{0.0265 \\ 0.0293 \\ 0.0230 \\ 0.0050} &
\makecell{0.6178 \\ 0.6913 \\ 0.6499 \\ 0.0999} &
\makecell{0.0107 \\ 0.0192 \\ 0.0499 \\ 0.0020} 
\\ \addlinespace[0.5mm]
\makecell{\textbf{\gls*{maskrcnn}} \\ \textbf{\gls*{resnet}-50}} & 
\makecell{\textbf{F-measure} \\ \textbf{Precision} \\ \textbf{Recall} \\ \textbf{\gls*{mae}}} &
\makecell{\textbf{\textcolor{blue}{0.6964}} \\ 0.7060 \\ 0.7646 \\
\textbf{\textcolor{blue}{0.0765}}} &
\makecell{0.0023 \\ 0.0034 \\ 0.0039 \\ 0.0012} &
\makecell{\textbf{\textcolor{blue}{0.8383}} \\ 0.8671 \\ 0.8137 \\
\textbf{\textcolor{blue}{0.0743}}} &
\makecell{0.0022 \\ 0.0028 \\ 0.0040 \\ 0.0010} &
\makecell{\textbf{\textcolor{blue}{0.7979}} \\ 0.8299 \\ 0.7781 \\
\textbf{\textcolor{blue}{0.0679}}} &
\makecell{0.0017 \\ 0.0030 \\ 0.0054 \\ 0.0004} &
\makecell{\textbf{\textcolor{blue}{0.7495}}\\ 0.7929 \\ 0.7356 \\ 0.0927} &
\makecell{0.0019 \\ 0.0016 \\ 0.0038 \\ 0.0007} &
\makecell{\textbf{\textcolor{blue}{0.7850}} \\ 0.8287 \\ 0.7722 \\
\textbf{\textcolor{blue}{0.0896}}} &
\makecell{0.0017 \\ 0.0025 \\ 0.0048 \\ 0.0008} &
\makecell{\textbf{\textcolor{blue}{0.8861}} \\ 0.9052 \\ 0.8644 \\
\textbf{\textcolor{blue}{0.0539}}} &
\makecell{0.0051 \\ 0.0057 \\ 0.0044 \\ 0.0018} &
\makecell{\textbf{\textcolor{blue}{0.7873}} \\ 0.8733 \\ 0.6813 \\
\textbf{\textcolor{blue}{0.0801}}} &
\makecell{0.0081 \\ 0.0037 \\ 0.0095 \\ 0.0023} &
\makecell{\textbf{\textcolor{blue}{0.6680}} \\ 0.6617 \\ 0.8014 \\
\textbf{\textcolor{blue}{0.0912}}} &
\makecell{0.0019 \\ 0.0026 \\ 0.0041 \\ 0.0007} 
\\ \midrule
\makecell{\textbf{\gls*{fcn}} \\ \textbf{\gls*{resnet}-101}} & 
\makecell{\textbf{F-measure} \\ \textbf{Precision} \\ \textbf{Recall} \\ \textbf{\gls*{mae}}} &
\makecell{0.4424 \\ 0.6410 \\ 0.3796 \\ 0.1078} &
\makecell{0.0327 \\ 0.0484 \\ 0.0463 \\ 0.0080} &
\makecell{0.7665 \\ 0.8970 \\ 0.6592 \\ 0.0949} &
\makecell{0.0176 \\ 0.0222 \\ 0.0308 \\ 0.0044} &
\makecell{0.7353 \\ 0.8658 \\ 0.6320 \\ 0.0757} &
\makecell{0.0216 \\ 0.0323 \\ 0.0358 \\ 0.0044} &
\makecell{0.6820 \\ 0.7694 \\ 0.6306 \\ 0.1080} &
\makecell{0.0287 \\ 0.0612 \\ 0.0503 \\ 0.0097} &
\makecell{0.6919 \\ 0.8174 \\ 0.6135 \\ 0.1070} &
\makecell{0.0156 \\ 0.0322 \\ 0.0287 \\ 0.0042} &
\makecell{0.5382 \\ 0.7879 \\ 0.4420 \\ 0.1527} &
\makecell{0.0520 \\ 0.0683 \\ 0.0640 \\ 0.0075} &
\makecell{0.3150 \\ 0.6299 \\ 0.2441 \\ 0.1862} &
\makecell{0.0444 \\ 0.1203 \\ 0.0637 \\ 0.0227} &
\makecell{0.5763 \\ 0.6581 \\ 0.6088 \\ 0.1021} &
\makecell{0.0191 \\ 0.0487 \\ 0.0439 \\ 0.0077} 
\\ \addlinespace[0.5mm]
\makecell{\textbf{\gls*{maskrcnn}} \\ \textbf{\gls*{resnet}-101}} & 
\makecell{\textbf{F-measure} \\ \textbf{Precision} \\ \textbf{Recall} \\ \textbf{\gls*{mae}}} &
\makecell{\textbf{\textcolor{red}{0.7072}} \\ 0.7166 \\ 0.7739 \\ \textbf{\textcolor{red}{0.0739}}}
&
\makecell{0.0033 \\ 0.0051 \\ 0.0028 \\ 0.0013} &
\makecell{\textbf{\textcolor{red}{0.8465}} \\ 0.8712 \\ 0.8258 \\ \textbf{\textcolor{red}{0.0694}}}
&
\makecell{0.0024 \\ 0.0028 \\ 0.0049 \\ 0.0007} &
\makecell{\textbf{\textcolor{red}{0.8066}} \\ 0.8384 \\ 0.7853 \\ \textbf{\textcolor{red}{0.0653}}}
&
\makecell{0.0023 \\ 0.0042 \\ 0.0044 \\ 0.0005} &
\makecell{\textbf{\textcolor{red}{0.7575}} \\ 0.7974 \\ 0.7442 \\ \textbf{\textcolor{red}{0.0886}}}
&
\makecell{0.0036 \\ 0.0044 \\ 0.0050 \\ 0.0013} &
\makecell{\textbf{\textcolor{red}{0.7897}} \\ 0.8347 \\ 0.7754 \\ \textbf{\textcolor{red}{0.0872}}}
&
\makecell{0.0005 \\ 0.0024 \\ 0.0036 \\ 0.0006} &
\makecell{\textbf{\textcolor{red}{0.9002}} \\ 0.9200 \\ 0.8710 \\ \textbf{\textcolor{red}{0.0480}}}
&
\makecell{0.0008 \\ 0.0020 \\ 0.0026 \\ 0.0004} &
\makecell{\textbf{\textcolor{red}{0.7911}} \\ 0.8745 \\ 0.6844 \\ \textbf{\textcolor{red}{0.0812}}}
&
\makecell{0.0039 \\ 0.0052 \\ 0.0097 \\ 0.0014} &
\makecell{\textbf{\textcolor{red}{0.6765}} \\ 0.6650 \\ 0.8227 \\ \textbf{\textcolor{red}{0.0880}}}
&
\makecell{0.0018 \\ 0.0030 \\ 0.0045 \\ 0.0007}
\\ \bottomrule
\end{tabular}
}
\label{tab:tests}
\end{table*}

As presented in Table~\ref{tab:tests}, the higher contribution of the \gls*{maskrcnn} was in the Recall, which indicates that the
\gls*{maskrcnn} impressively reduced the false negative error, as shown in Fig.~\ref{fig:segmentation_example}. With the
\gls*{resnet}-50 backbone, the \gls*{maskrcnn} overcame the \gls*{fcn} F-measure by 37.9\% in the SED2, 24.51\% in the SED1,
and 19.65\% in the DUT-OMRON datasets in the best cases and overcame the \gls*{fcn} F-measure by 0.07\% in the ICOSEG dataset in the
worst case. Also, with exception of the ICOSEG dataset, the \gls*{maskrcnn} achieved lower \glspl*{mae} when compared with the
\gls*{fcn}, both with \gls*{resnet}-50 backbone. With the \gls*{resnet}-101 backbone, the \gls*{maskrcnn} overcame the \gls*{fcn}
F-measure by 7.13\% in the HKU-IS dataset in the worst case. However, the \gls*{maskrcnn} overcame the \gls*{fcn} F-measure by
47.61\% in the SED2, 36.20\% in the SED1, and 26.48\% in the DUT-OMRON datasets, both with \gls*{resnet}-101 backbone. Also, the
\gls*{maskrcnn} achieved lower \glspl*{mae} in all datasets when compared with the \gls*{fcn}.

\section{Conclusion}
In this work, we compared two versions of \glspl*{fcn} with two versions of \glspl*{maskrcnn} through extensive experiments on eight
public datasets using four metrics (F-measure, Precision, Recall, and \gls*{mae}). We have shown that the standard \glspl*{maskrcnn}
with \glspl*{resnet} backbones outperforms the standard \glspl*{fcn}, with the same backbones, in the \gls*{sod} problem. The
\gls*{rpn} attached before the segmentation module in the \gls*{maskrcnn} improved the segmentation results by significantly
decreasing the false negative error. Although, the \glspl*{maskrcnn} evaluated in this work are outperformed by state-of-art
results~\cite{Liu2019PoolSal} based on enhanced \glspl*{fcn}, further research enhancing the \glspl*{maskrcnn} in a similar way could
provide competitive results. Our findings endorse that the \gls*{maskrcnn} is a promising alternative to solve the \gls*{sod}
problem.


\section*{ACKNOWLEDGMENT}
The authors would like to thank the Coordination for the Improvement of Higher Education Personnel
 (CAPES) for the Masters scholarship. We gratefully acknowledge the founders of the publicly
 available datasets and the support of NVIDIA Corporation with the donation of the GPUs used
 for this research.


\balance
\bibliographystyle{IEEEtran}
\bibliography{main}

\end{document}